\documentclass[wcp]{jmlr}

\usepackage{longtable}% for long tables

\usepackage{booktabs}
\usepackage{latexsym}
\usepackage{amsmath}
\usepackage{makecell}
\usepackage{xcolor}
\usepackage{graphicx}
\usepackage{multirow}
\usepackage{comment}
\usepackage{url}
\usepackage{xcolor}
\usepackage{soul}
\usepackage{tikz}
\newcommand{\hlc}[2][yellow]{{%
    \colorlet{foo}{#1}%
    \sethlcolor{foo}\hl{#2}}%
}

\definecolor{ls}{rgb}{0.99,0.95,0.85}
\definecolor{sl}{rgb}{0.88,0.84,0.92}

\newcolumntype{Y}{>{\centering\let\newline\\\arraybackslash\hspace{0pt}}X}

\usepackage{paralist}

\jmlrvolume{101}
\jmlryear{2019}
\jmlrworkshop{ACML 2019}

\title[Cross Length Transfer]{Text Length Adaptation in Sentiment Classification}

\author{
 \Name{Reinald Kim Amplayo}\thanks{Most work done when both first and second authors were at Yonsei University.} \Email{reinald.kim@ed.ac.uk}\\
 \addr University of Edinburgh, UK
 \AND
 \Name{Seonjae Lim} \Email{sun.lim@samsung.com}\\
 \addr Samsung Electronics, South Korea
 \AND
 \Name{Seung-won Hwang} \Email{seungwonh@yonsei.ac.kr}\\
 \addr Yonsei University, South Korea
}

\editors{Wee Sun Lee and Taiji Suzuki}

\makeatletter
\newcommand{\thickhline}{%
    \noalign {\ifnum 0=`}\fi \hrule height 1pt
    \futurelet \reserved@a \@xhline
}
\makeatother

\begin{document}

\maketitle

\begin{abstract}
  Can a text classifier generalize well for datasets where the text length is different? For example, when short reviews are sentiment-labeled, can these transfer to predict the sentiment of long reviews (i.e., short to long transfer), or vice versa? While unsupervised transfer learning has been well-studied for cross domain/lingual transfer tasks, \textbf{Cross Length Transfer} (CLT) has not yet been explored. One reason is the assumption that length difference is trivially transferable in classification. We show that it is not, because short/long texts differ in context richness and word intensity. We devise new benchmark datasets from diverse domains and languages, and show that existing models from similar tasks cannot deal with the unique challenge of transferring across text lengths. We introduce a strong baseline model called \textsc{BaggedCNN} that treats long texts as bags containing short texts. We propose a state-of-the-art CLT model called \textbf{Le}ngth \textbf{Tra}nsfer \textbf{Net}work\textbf{s} (\textsc{LeTraNets}) that introduces a two-way encoding scheme for short and long texts using multiple training mechanisms. We test our models and find that existing models perform worse than the \textsc{BaggedCNN} baseline, while \textsc{LeTraNets} outperforms all models.
\end{abstract}

\section{Introduction}

\begin{figure}[t]
    \centering
    \begin{subfigure}
        \centering
        \includegraphics[width=0.7\textwidth]{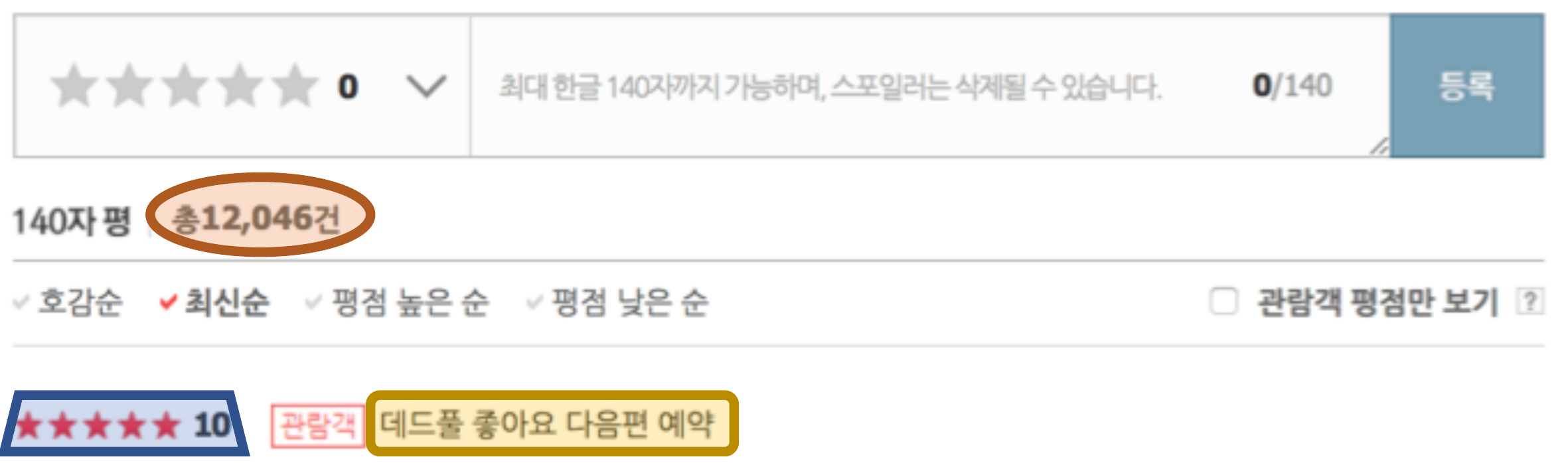}
        %\caption{Short Review Channel}
        %\label{fig:ex_short}
    \end{subfigure}
    \begin{subfigure}
        \centering
        \includegraphics[width=0.7\textwidth]{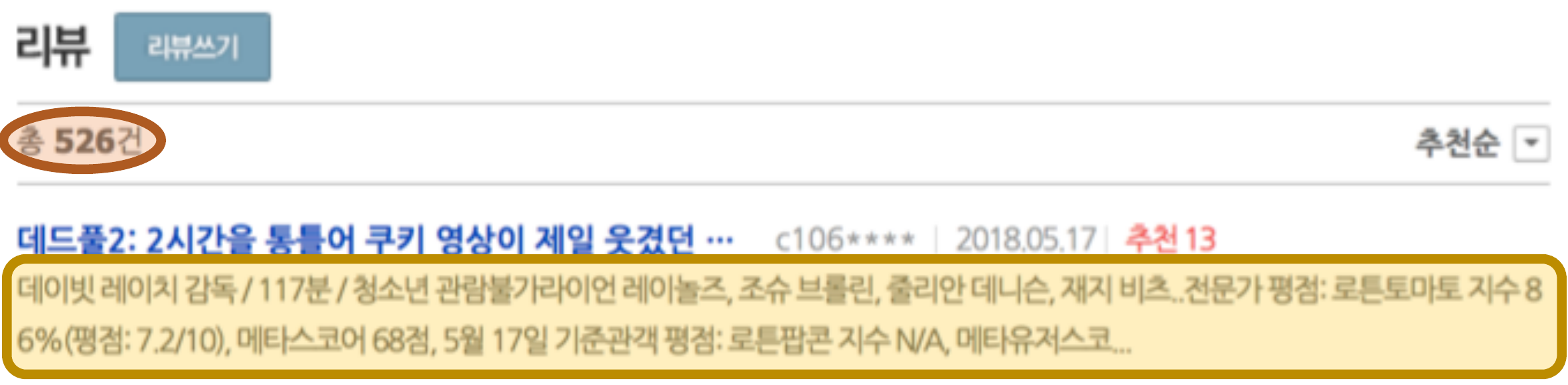}
        %\caption{Long Review Channel}
        %\label{fig:ex_long}
    \end{subfigure}
    \caption{Short review (a) and long review (b) channels for users to write reviews in Naver Movie website. Highlighted areas emphasize the difference in \textcolor{red}{number of reviews}, \textcolor{orange}{review text length}, and \textcolor{blue}{sentiment label availability}.}
    \label{fig:ex}
\end{figure}

Text classification can be categorized according to the text length of the data, from sentence-level classification \citep{kim2014convolutional} to document-level classification \citep{manevitz2001one,yang2016hierarchical}.
One kind of such task is sentiment classification \citep{pang2002thumbs}, a subtask of sentiment analysis \citep{pang2008opinion,liu2012sentiment} where we are to predict the sentiment/rating given a review written by a user.
In some domains, the length of these reviews varies widely. For example,
well-known review websites in East Asia such as Naver Movies\footnote{\url{https://movie.naver.com/}} and Douban Movies\footnote{\url{https://movie.douban.com/}} provide two channels for users to write reviews, depending on their preferred length. Figure \ref{fig:ex} shows the review channels provided in Naver Movies.

The first channel is a \textit{short review channel}, which contains large amounts of reviews, and enforces users to write short reviews accompanied by rating labels.
Although labeled, these reviews lack expressiveness to extract useful information.
In contrast, the second channel
is a \textit{long review channel}, which contains few long and detailed reviews
with descriptions of different aspects about the product/service.
Despite being more expressive, long reviews are often not accompanied by sentiment labels, which most supervised sentiment classification models would require.

We study the ``transferability'' from one review channel to the other. That is, we try to answer whether a text classifier trained on a dataset with length $\alpha$ can predict a text with length $\beta$, where $\alpha$ and $\beta$ differ by a large margin (e.g., sentences versus paragraphs).
This is an important question because there are scenarios where we may have better and more sufficient labeled text data, but we want to classify unlabeled texts with different length.
For long to short transfer, more expressive long reviews can be leveraged for training a sentiment classifier for short and context-sparse reviews.
For short to long transfer, large amounts of short reviews can be used as supervision to train a classifier for long reviews.

%To answer the question on the transferability between texts of different length, 
To motivate the non-triviality of such transfer,
we train an out-channel (OC) classifier that uses short texts to predict long texts, and an in-channel (IC) classifier that uses long texts on both training and prediction. We also experiment conversely. We use three kinds of classifiers: bag-of-words (BoW), convolutional neural networks \cite[CNN;][]{kim2014convolutional}, and BERT multilingual \citep{devlin2018bert}.
We calculate the transfer loss \cite[TL;][]{glorot2011domain}, which is the difference between the out-channel and in-channel classifier errors (i.e.,~$\text{TL}=0$ means trivially transferable).
Table \ref{tab:prior} shows that, though using a better inductive bias such as CNN and BERT seems to slightly lower TL, it remains significantly high, consistently suggesting that length transfer is non-trivial.

% Table generated by Excel2LaTeX from sheet 'Sheet6'
\begin{table}[t]
  \small
  \centering
    \begin{tabular}{ccccccc}
    \thickhline
    \multirow{1}[4]{*}{} & \multicolumn{2}{c}{\textsc{Mov\_en}} & \multicolumn{2}{c}{\textsc{Res\_en}} & \multicolumn{2}{c}{\textsc{Mov\_ko}} \\
  & L\textgreater S  & S\textgreater L  & L\textgreater S  & S\textgreater L  & L\textgreater S  & S\textgreater L \\
    \hline
    BoW & {4.2}   &  9.8 & {5.8}  & {9.8}   & {17.5}  & {13.0} \\
    CNN & 5.5 & 5.7 & 5.8 & 11.8 & 6.3 & 14.5 \\
    BERT & 5.3 & 5.7 & 4.5 & 8.5 & 3.0 & 10.0 \\
    \hline
    \textsc{LeTraNets} & -1.2 & 0.5 & 3.2 & -1.5 & 0.6 & 4.5 \\
    \thickhline
    \end{tabular}%
  \caption{Transfer loss (TL) of different classifiers for each transfer task for all datasets described in Section \ref{sec:datasets}.}
  \label{tab:prior}%
\end{table}%

Our first contribution is thus to define a new task called Cross Length Transfer (CLT). CLT is a task similar to Cross Domain \citep{blitzer2007biographies} and Cross Lingual Transfer \citep{mihalcea2007learning} where the difference between the source and target texts is the \textit{text length}, having non-trivial influence that is shown in Table \ref{tab:prior}.
Our second contribution is to show that models from similar tasks (e.g. Cross Domain Transfer and Multiple Instance Learning) are not effective for CLT and even yield negative transfer, as we elaborate in Section \ref{sec:competing} and empirically show in Section \ref{sec:experiments}.
Finally, we present two new models specifically for CLT: a strong baseline called \textsc{BaggedCNN} that treats long texts as bags containing short texts, and a state-of-the-art CLT model called Length Transfer Networks (\textsc{LeTraNets}). \textsc{LeTraNets} enables a two-way encoding scheme using multiple training mechanims, and accepts both short and long text inputs, where one such input is created artificially through concatenation or segmentation.
Table \ref{tab:prior} shows that \textsc{LeTraNets} has the best transfer loss, and sometimes perform better than in-channel classifier (when TL is less than zero).
%by introducing pseudo-inputs to enable a two-way encoding scheme that captures
%both stand-alone and segment-level text features.

We test our models using the multiple benchmark datasets we gathered and show that models from other tasks perform worse than our proposed strong baseline and that \textsc{LeTraNets} performs the best among all models. To the best of our knowledge, we are the first to study CLT.

\section{Cross Length Transfer}

Cross Length Transfer (CLT) is an unsupervised transfer learning task in which the setting is that the sampling distributions of the training and test data are different because the texts lengths are different (e.g., sentences and paragraphs).

Formally, we suppose two sets of texts: a source set $\mathcal{S}$ in which we have labels, and a target set $\mathcal{T}$ in which we want to predict the labels. Moreover, we know that the text length distributions of $\mathcal{S}$ and $\mathcal{T}$ are different, such that an equality case exists as $|\mathcal{S}| = r |\mathcal{T}|$, where $|\mathcal{X}|$ is the mean length of the set $\mathcal{X}$, and $r \neq 1$ is a non-negative rate of difference between two mean lengths.
There are two subtasks: \textbf{long to short transfer} where $r>1$ and thus $\mathcal{S}$ contains longer texts, and \textbf{short to long transfer} where $r<1$ and thus $\mathcal{S}$ contains shorter texts.
A CLT model should effectively learn to predict the labels of $\mathcal{T}$, on both scenarios.

A concrete and simple example is when $\mathcal{S}$ contains labeled sentence reviews and $\mathcal{T}$ contains unlabeled paragraph reviews. A CLT model uses $\mathcal{S}$ for training to effectively predict labels of reviews in $\mathcal{T}$. Also, the same CLT model should be able to do effective prediction vice versa, i.e., when $\mathcal{S}$ are paragraph reviews and $\mathcal{T}$ are sentence reviews.

Previous unsupervised transfer learning tasks, i.e. Cross Domain Transfer \citep{blitzer2007biographies} and Cross Lingual Transfer \citep{wan2009co}, are similar to CLT but have concrete differences. Generally, the goal of these tasks is to map semantic domains, contextually or linguistically, of both $\mathcal{S}$ and $\mathcal{T}$ into a shared space, by aligning the vocabulary \citep{pan2010cross,shi2018learning},
expanding domain-specific lexicons \citep{qiu2009expanding,he2011automatically},
generating labeled samples \citep{yu2016learning}, and 
learning to indiscriminate between domains \citep{chen2016adversarial,liu2018learning}.
These methods are generally symmetric; i.e., even when $\mathcal{S}$ and $\mathcal{T}$ interchange, the same method can be applied easily.

However, in CLT, both $\mathcal{S}$ and $\mathcal{T}$ are already in the same contextual and linguistic domains, thus previous methods would not work. Also, CLT brings two new challenges 
against devising a symmetric model. 
First, texts with different context richness may have different properties they focus on: hierarchical structures \citep{yang2016hierarchical} may be more important for document-level reviews while finding lexical/phrasal cues \citep{kim2014convolutional} may be more important for sentence-level reviews. Second, words on texts with different lengths may have different semantic intensity. For example, ``\textit{good}'' may have a very high positive sentiment intensity on short texts, and a relatively low positive sentiment intensity on long ones.

\subsection{Benchmark Datasets}
\label{sec:datasets}

\begin{table}[t]
  \centering
    \begin{tabular}{lccccc}
    \thickhline
    \multicolumn{1}{c}{Dataset} & Length & $\#$Train & $\#$Test & $\#$Unlabeled & Words/Instance \\
    \hline
    \multirow{2}[0]{*}{\textsc{Mov\_en}} & short & 1600  & 400   & 8000  & 22 \\
          & long  & 1600  & 400   & 8000  & 372 \\
    \multirow{2}[0]{*}{\textsc{Res\_en}} & short & 1600  & 400   & 8000  & 22  \\
          & long  & 1600  & 400   & 8000  & 249\\
    \multirow{2}[0]{*}{\textsc{Mov\_ko}} & short & 1600  & 400   & 4000  & 12 \\
          & long  & 1600  & 400   & 0     & 57\\
    \thickhline
    \end{tabular}%
  \caption{Dataset statistics.}
  \label{tab:data}%
\end{table}%

We provide three pairs of short/long datasets from different domains (movies and restaurants) and from different languages (English and Korean) suitable for the task: \textsc{Mov\_en}, \textsc{Res\_en}, and \textsc{Mov\_ko}.
%\footnote{We define the short reviews as sentence-level, and the long reviews as multiple sentence document-level reviews.}
Most of the datasets are from previous literature
and are gathered differently
The \textsc{Mov\_en} datasets are gathered from different websites; the short dataset consists of hand-picked sentences by \citet{pang2005seeing} from document-level reviews from the Rotten Tomatoes website, while the long dataset consists of reviews from the IMDB website obtained by \citet{tang2015learning}.
The \textsc{Res\_en} dataset consists of reviews from Yelp, where the short dataset consists of reviews with character lengths less than 140 from \citet{amplayo2017aspect}, while reviews in the long dataset are gathered from \citet{tang2015learning}.
We also share new short/long datasets \textsc{Mov\_ko}, which are gathered from two different channels, as shown in Figure \ref{fig:ex}, available in Naver Movies.

Unlike previous datasets \citep{blitzer2007biographies,glorot2011domain} where they used polarity/binary (e.g., positive or negative) labels as classes, we also provide fine-grained classes, with five classes of different sentiment intensities (e.g., 1 is strong negative, 5 is strong positive), for \textsc{Res\_en} and \textsc{Mov\_ko}.
Following the Cross Domain Transfer setting \citep{blitzer2007biographies,ziser2017neural,plank2018strong}, we limit the size of the dataset to be small-scale to focus on the main task at hand. This ensures that models focus on the transfer task, and decrease the influence of other factors that can be found when using larger datasets.
Finally, following \citet{glorot2011domain}, we provide additional unlabeled data for those models that need them \citep{blitzer2007biographies,ziser2017neural}, except for the long dataset of \textsc{Mov\_ko}, where the labeled reviews are very limited.
We show the dataset statistics in Table \ref{tab:data}, and share the datasets here: \url{https://github.com/rktamplayo/LeTraNets}.

\subsection{Possible Existing Solutions}
\label{sec:competing}

\paragraph{Cross Domain Transfer (CDT)}

CDT offers models that effectively transfer domain-independent features from two different domains. The most popular non-neural CDT model is Structural Correspondence Learning \cite[\textsc{SCL;}][]{blitzer2007biographies}, a method that identifies feature correspondence from different domains using pivot features. A recent neuralized extension is Neural SCL \cite[\textsc{NeuSCL;}][]{ziser2017neural}, in which an autoencoder module is integrated to SCL. The CDT literature is vast, and we refer the readers to \citet{pan2012transfer} and \citet{tan2018survey} for overviews. Although these models may see improvements due to a possible difference in vocabulary (especially when the review channels are different), these improvements may be marginal since the domain of the datasets is the same. %This argument is empirically shown in our experiments in Section \ref{sec:experiments}.

\paragraph{Multiple Instance Learning (MIL)}

MIL is a task where given the labels of a bag of multiple instances, we are to label the individual instances \citep{zhou2009multi}. In the text classification domain, MIL is often devised as segment-level classification \citep{kotzias2015group,angelidis2018multiple}, where documents are bags and sentences in the documents are segments. The most recent MIL model is the Multiple Instance Learning Network \cite[\textsc{MILNet};][]{angelidis2018multiple}, where they used attention-based polarity scoring to identify segment labels. MIL models can be used in \textit{long to short transfer}, where we assume that segment labels in long texts can be used to label short reviews. However, they (a) assume that segments from long data, which rely on inter-sentence semantics, are comparable to self-contained short texts, and (b) are ineffective on short to long transfer because it needs multiple sentences to train components of the model for document-level classification.

\paragraph{Weak Supervision}

A simple yet possible solution for \textit{short to long transfer} is a three-step approach where we (1) cluster the short texts into several long texts, (2) infer the class labels of the clusters, and (3) use the labeled clusters as weak supervision to create a classifier. Micro Aspect Sentiment Model \cite[\textsc{MASM};][]{amplayo2017aspect} does (1) and (2) automatically. For (3), we can train a classifier such as CNNs \citep{kim2014convolutional} to predict labels of long texts. One critical issue of this solution is that since both clustering and class labels are inferred, there is a high chance that at least one of them is incorrect. This thus creates compounding errors that decrease the performance of the model.

\section{Our Models}

\subsection{\textsc{BaggedCNN}: A Strong Baseline}

We present \textsc{BaggedCNN}, a simple yet strong baseline to the CLT task. \textsc{BaggedCNN} is a model derived from \textsc{MILNet} \citep{angelidis2018multiple}. \textsc{MILNet} uses CNN to encode segments, BiGRU \citep{bahdanau2014neural} to calculate attention weights, and gated polarity to calculate document-level probabilities. We refer the readers to the original paper for more details. We improve using two key modifications: (a) removing the sequential connections (i.e., BiGRU) between segments, and (b) using a single classifier for both the segments and full document.

For each document divided into segments $D=\{S_i\}$, \textsc{BaggedCNN} starts by encoding the segments using a \textsc{CNN} classifier called $\text{CNN}_{bag}$. Then, we pool the segment encodings into one vector using attention mechanism. Finally, we use a logistic regression classifier that can be used to classify either the segments or the document. This is possible since the vectors of both segments and document are in the same vector space:
\begin{align*}
    s_i &= \text{CNN}_{bag}(S_i) \\
    a_{s_i} &= \text{softmax}(v^\top \text{tanh}(W_a s_i + b_a)) \\
    d &= \sum_i s_i * a_{s_i} \\
    y_d &= \text{softmax}(W_c d + b_c) \\
    y_{s_i} &= \text{softmax}(W_c s_i + b_c)
\end{align*}

The model is trained differently depending on the transfer task: For \textit{long to short transfer}, we minimize the cross-entropy loss between the actual and predicted class of the document $\mathcal{L}_d$. For \textit{short to long transfer}, we minimize the mean cross entropy loss between the actual and predicted class of the segments $\sum \mathcal{L}_{s_i}/n, 1\leq i \leq n$. Note that \textsc{BaggedCNN} is reduced to a model where average pooling is done instead of the attention mechanism. At test time, we use $y_d$ for classification.

While it has been shown that removing the sequential structure in the document level (i.e., BiGRU in the case of \textsc{MILNet}) decreases the performance of the document classifier \citep{tang2015document,yang2016hierarchical}, we argue that this removal is effective on the CLT task because of inter-segment independence. That is, sentences in the document are treated similar to short texts. We also show in our experiments that \textsc{BaggedCNN} performs better than \textsc{MILNet}.

However, \textsc{BaggedCNN} still fails to consider two things. First, while the model relaxes the strong assumption on similarity between segments and short texts, by removing the sequential connections, most segments cannot be treated as stand-alone short texts. For example, the segment ``\textit{Yet it is salty.}'' is not a stand-alone short review. Second, when doing short to long transfer, the input short text is just one segment, thus the model is reduced into a weaker hierarchical CNN classifier.

\subsection{\textsc{LeTraNets}: Length Transfer Networks}

\begin{figure*}
    \centering
    \includegraphics[width=0.9\textwidth]{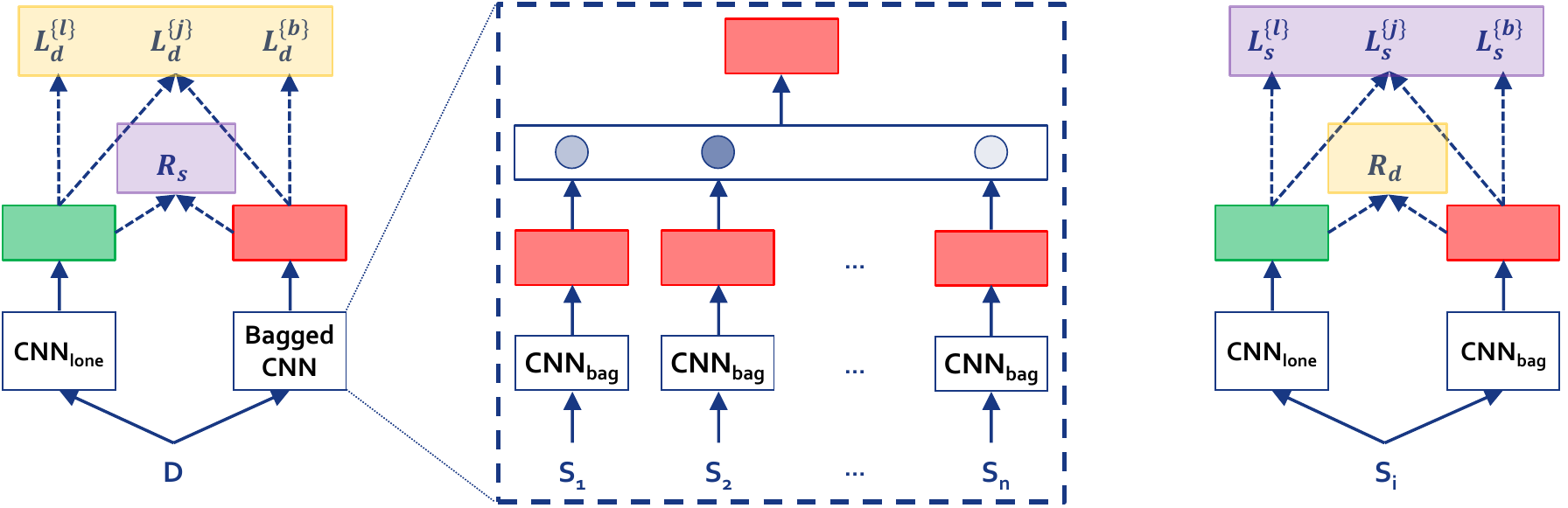}
    \caption{Full architecture of \textsc{LeTraNets}. $\mathcal{L}^{\{a\}}_x$ refers to cross entropy loss of classifier $a$ ($l$: $p(y^{\{l\}})$, $b$: $p(y^{\{b\}})$, $j$: $p(y^{\{j\}})$) using text $x$ ($s$: short text, $d$: long text). $\mathcal{R}_x$ is the prediction regularization. The highlighted boxes refer to the \hlc[ls]{long to short} and \hlc[sl]{short to long} transfer-specific configurations. We share our implementation here: \url{https://github.com/rktamplayo/LeTraNets}.}
    \label{fig:letranets}
\end{figure*}

We improve \textsc{BaggedCNN} by proposing a model called Length Transfer Networks (\textsc{LeTraNets}), as shown in Figure \ref{fig:letranets}.
\textsc{LeTraNets} is composed of two classifiers: a stand-alone CNN classifier with text encoder $\text{CNN}_{lone}$, and \textsc{BaggedCNN}, which includes a segment-level text encoder $\text{CNN}_{bag}$. The $\text{CNN}_{lone}$ encoder is used to capture holistic textual features, while the $\text{CNN}_{bag}$ encoder is used to capture segment-level textual features, assuming there is a bigger text that owns the segments.

For each data instance, \textsc{LeTraNets} accepts two kinds of inputs: a long text $D={w_d}$ and a set of short texts $S={{w_{s_0}},...,{w_{s_n}}}$.
However, the task setting only provides either one of long texts or short texts as input. We thus create pseudo-texts from the available text data through the following methods. In the long to short transfer task, we use segments in long texts as pseudo-short texts, as used in \textsc{BaggedCNN}. In the short to long transfer task, we concatenate a random number of short texts to create pseudo-long texts. The latter amounts to a possibly infinite number of long texts we can use for training.

The short texts are encoded by both $\text{CNN}_{lone}$ and $\text{CNN}_{bag}$ as $s^{\{l\}}_i$ and $s^{\{b\}}_i$. The long texts are encoded using both $\text{CNN}_{lone}$ and \textsc{BaggedCNN} as $d^{\{l\}}$ and $d^{\{b\}}$:
\begin{align}
    \tiny
    s^{\{l\}}_i &= \text{CNN}_{lone}(S_i) \nonumber \\ 
    s^{\{b\}}_i &= \text{CNN}_{bag}(S_i) \nonumber \\
    d^{\{l\}} &= \text{CNN}_{lone}(D) \nonumber \\
    d^{\{b\}} &= \textsc{BaggedCNN}(D) \nonumber
\end{align}

The encoded long text vectors $d^{\{l\}}$ and $d^{\{b\}}$ and short text vectors $s^{\{b\}}_i$ and $s^{\{l\}}_i$ are used to classify their labels using softmax classifiers specific to the CNN encoders:
\begin{align}
    \small
    p(y^{\{l\}}_{d}) &= \text{softmax}(W^{\{l\}}_c d^{\{l\}} + b^{\{l\}}_c) \nonumber \\
    p(y^{\{b\}}_{d}) &= \text{softmax}(W^{\{b\}}_c d^{\{b\}} + b^{\{b\}}_c) \nonumber \\
    p(y^{\{l\}}_{s_i}) &= \text{softmax}(W^{\{l\}}_c s^{\{l\}}_i + b^{\{l\}}_c) \nonumber \\
    p(y^{\{b\}}_{s_i}) &= \text{softmax}(W^{\{b\}}_c s^{\{b\}}_i + b^{\{b\}}_c) \nonumber
\end{align}

\paragraph{Training Mechanisms}

There are two main issues when training the model in the CLT setting. First, both the stand-alone CNN classifier and \textsc{BaggedCNN} are disconnected, acting as two individual classifiers. Second, the model needs both labels for both short and long text data, but we are only given labels for one kind of data during training for each transfer setting. Solving the second issue is crucial for short to long transfer, as we cannot train the full model if we do not have labels for long data. To this end, we use three training mechanisms below that help mitigate these issues.

We connect them on different levels. In the word-level, we use the same word embedding space for both classifiers. Beyond word-level, we use a training mechanism called \textbf{Joint Training (JT)}. This concatenates the encoded text vectors, and creates another logistic regression classifier for the concatenated vector. This creates a connection between classifiers at the classification-level.
\begin{align*}
    p(y^{\{j\}}_{d}) &= \text{softmax}(W^{\{j\}}_c [d^{\{l\}};d^{\{b\}}] + b^{\{j\}}_c) \\
    p(y^{\{j\}}_{s_i}) &= \text{softmax}(W^{\{j\}}_c [s^{\{l\}}_i;s^{\{b\}}_i] + b^{\{j\}}_c)
\end{align*}

Beyond word-level, we introduce \textbf{Prediction Regularization (PR)} mechanism to train encoders with no labels. This regularizes the predictions of a weaker classifier based on the predictions of a stronger classifier. We consider \textsc{BaggedCNN} as the stronger classifier for long to short transfer, and $\text{CNN}_{lone}$ as the stronger classifier for short to long transfer. We use Kullback-Leibler divergence as the regularization function.
\begin{align*}
    \mathcal{R}_d &= \sum_i \mathcal{D}_{KL}(y^{b}_{s_i} || y^{l}_{s_i}) \\
    \mathcal{R}_s &= \mathcal{D}_{KL}(y^{l}_{d} || y^{b}_{d})
\end{align*}

Finally, using the PR mechanism directly might not work because predictions from the stronger classifier may not be optimized yet. Hence, we use \textbf{Stepwise Pretraining (SP)} mechanism to pretrain specific parts of the model in a step-by-step fashion. First, we pretrain the stronger classifier, then the weaker classifier with PR mechanism, and finally the classifier of the JT mechanism. After pretraining, we train the full model.

The training configurations are different depending on the transfer task, which is also shown in Figure \ref{fig:letranets}. For \textit{long to short transfer}, we use $p(y^{\{j\}}_{d})$ for the JT mechanism and $R_d$ for the PR mechanism. For \textit{short to long transfer}, we use $p(y^{\{j\}}_{s_i})$ for the JT mechanism and $R_s$ for the PR mechanism.

The final training objective is to minimize the loss function, depending on the text length:
\begin{align*}
\mathcal{L}_d &= \mathcal{L}^{\{l\}}_d + \mathcal{L}^{\{b\}}_d + \mathcal{L}^{\{j\}}_d + \lambda \mathcal{R}_d \\
\mathcal{L}_s &= \sum_i (\mathcal{L}^{\{l\}}_{s_i} + \mathcal{L}^{\{b\}}_{s_i} + \mathcal{L}^{\{j\}}_{s_i}) / n + \lambda \mathcal{R}_s
\end{align*}
where $\mathcal{L}^{\{a\}}_x$ is the cross-entropy loss between the actual and predicted values of the classifier $p(y^{\{a\}}_x)$, and $\lambda$ is tuned using a development set. At test time, we use $p(y^{\{j\}}_{d})$ and $p(y^{\{j\}}_{s_i})$ to classify the sentiment for long to short and short to long transfer, respectively.

\section{Experiments}
\label{sec:experiments}

% Table generated by Excel2LaTeX from sheet 'Sheet5'
\begin{table*}[t]
  \footnotesize
  \centering
    \begin{tabular}{@{}l@{\hskip 3em}c@{\hskip 3em}ccc@{\hskip 3em}ccc@{}}
    \thickhline
    & \textsc{Mov\_en} & \multicolumn{3}{c@{\hskip 5em}}{\textsc{Res\_en}} & \multicolumn{3}{c@{\hskip 2em}}{\textsc{Mov\_ko}} \\
     \multicolumn{1}{c}{Model}  & \textsc{P\_Acc} & \textsc{P\_Acc} & \textsc{F\_Acc} & \textsc{F\_RMSE}  & \textsc{P\_Acc} & \textsc{F\_Acc} & \textsc{F\_RMSE} \\
    \thickhline
    \multicolumn{8}{c}{\textit{Long to Short Transfer Task}} \\
    \hline
    \textsc{CNN}   & 0.728 & 0.750 & 0.388 & 1.440 & 0.600 & 0.228 & 1.678 \\
    \textsc{CNN}x2   & 0.735 & 0.756 & 0.401 & 1.431 & 0.603 & 0.228 & 1.674 \\
    \hline
    \textsc{SCL}   & \textcolor[rgb]{ 1,  0,  0}{0.640} & \textcolor[rgb]{ 1,  0,  0}{0.685} & \textcolor[rgb]{ 1,  0,  0}{0.340} & \textcolor[rgb]{ 1,  0,  0}{1.857} & \textcolor[rgb]{ 1,  0,  0}{0.438} & \textcolor[rgb]{ 1,  0,  0}{0.208} & \textcolor[rgb]{ 1,  0,  0}{1.754} \\
    \textsc{NeuSCL}  & \textcolor[rgb]{ 1,  0,  0}{0.645} & \textcolor[rgb]{ 1,  0,  0}{0.665} & \textcolor[rgb]{ 1,  0,  0}{0.373} & \textcolor[rgb]{ 1,  0,  0}{1.765} & \textcolor[rgb]{ 1,  0,  0}{0.428} & \textcolor[rgb]{ 1,  0,  0}{0.203} & \textcolor[rgb]{ 1,  0,  0}{1.709} \\
    \textsc{SCL+CNN} & \textcolor[rgb]{ 1,  0,  0}{0.710} & 0.788 & 0.390 & 1.280 & \textcolor[rgb]{ 1,  0,  0}{0.578} & 0.230 & \textcolor[rgb]{ 1,  0,  0}{1.811} \\
    \textsc{NeuSCL+CNN} & 0.730 & 0.798 & 0.425 & \textcolor[rgb]{ 1,  0,  0}{1.444} & 0.600 & 0.230 & \textcolor[rgb]{ 1,  0,  0}{1.683} \\
    \textsc{MILNet} & 0.757 & 0.785 & 0.400 & 1.426 & 0.620 & \textcolor[rgb]{ 1,  0,  0}{0.212} & \textcolor[rgb]{ 1,  0,  0}{1.685} \\
    \hline
    \textsc{BaggedCNN} & 0.767 & 0.765 & 0.468 & 1.190 & 0.625 & 0.245 & 1.614 \\
    \textsc{\textbf{LeTraNets}} & \textbf{0.795*} & \textbf{0.863*} & \textbf{0.502*} & \textbf{1.081*} & \textbf{0.652} & \textbf{0.265} & \textbf{1.591} \\
    \thickhline
    \multicolumn{8}{c}{\textit{Short to Long Transfer Task}} \\
    \hline
    \textsc{CNN}   & 0.758 & 0.780 & 0.390 & 1.374 & 0.552 & 0.453 & 0.973 \\
    \textsc{CNN}x2   & 0.763 & 0.784 & 0.396 & 1.372 & 0.559 & 0.460 & 0.960 \\
    \hline
    \textsc{SCL}   & \textcolor[rgb]{ 1,  0,  0}{0.700} & \textcolor[rgb]{ 1,  0,  0}{0.738} & \textcolor[rgb]{ 1,  0,  0}{0.343} & 1.361 & \textcolor[rgb]{ 1,  0,  0}{0.488} & \textcolor[rgb]{ 1,  0,  0}{0.220} & \textcolor[rgb]{ 1,  0,  0}{1.568} \\
    \textsc{NeuSCL}  & \textcolor[rgb]{ 1,  0,  0}{0.725} & \textcolor[rgb]{ 1,  0,  0}{0.725} & \textcolor[rgb]{ 1,  0,  0}{0.385} & 1.251 & \textcolor[rgb]{ 1,  0,  0}{0.480} & \textcolor[rgb]{ 1,  0,  0}{0.245} & \textcolor[rgb]{ 1,  0,  0}{1.546} \\
    \textsc{SCL+CNN} & \textcolor[rgb]{ 1,  0,  0}{0.720} & \textcolor[rgb]{ 1,  0,  0}{0.753} & 0.403 & 1.190 & 0.555 & \textcolor[rgb]{ 1,  0,  0}{0.345} & \textcolor[rgb]{ 1,  0,  0}{1.250} \\
    \textsc{NeuSCL+CNN} & 0.783 & \textcolor[rgb]{ 1,  0,  0}{0.778} & 0.448 & 1.366 & 0.568 & 0.460 & 0.885 \\
    \textsc{MASM+CNN} & \textcolor[rgb]{ 1,  0,  0}{0.570} & \textcolor[rgb]{ 1,  0,  0}{0.600} & \textcolor[rgb]{ 1,  0,  0}{0.235} & \textcolor[rgb]{ 1,  0,  0}{2.322} & \textcolor[rgb]{ 1,  0,  0}{0.543} & \textcolor[rgb]{ 1,  0,  0}{0.045} & \textcolor[rgb]{ 1,  0,  0}{1.658} \\
    \hline
    \textsc{BaggedCNN} & 0.792 & 0.838 & 0.438 & 1.224 & 0.585 & 0.465 & 0.865 \\
    \textsc{\textbf{LeTraNets}} & \textbf{0.810} & \textbf{0.858} & \textbf{0.478*} & \textbf{1.138*} & \textbf{0.625*} & \textbf{0.493} & \textbf{0.802} \\
    \thickhline
    \end{tabular}%
  \caption{Accuracy and RMSE of competing models on polarity (\textsc{P\_Acc}) and fine-grained (\textsc{F\_Acc} and \textsc{F\_RMSE}) datasets. Items in \textcolor{red}{red} are performances worse than the no-transfer CNN baseline. 
  An asterisk (*) indicates that LeTraNets is significantly better than the second best model ($p < 0.05$) .}
  \label{tab:clt}%
\end{table*}%

\paragraph{Experimental Settings}

The dimensions of word vectors are set to 300. We use pre-trained GloVe embeddings\footnote{\url{https://nlp.stanford.edu/projects/glove/}} 
\citep{pennington2014glove} to initialize our English word vectors, and pre-trained FastText embeddings\footnote{\url{https://fasttext.cc/}} 
\citep{grave2018learning} to initialize our Korean word vectors. For all CNNs, we set $h=3,4,5$, each with 100 feature maps, following \citep{kim2014convolutional}. We use dropout \citep{srivastava2014dropout} on all non-linear connections with a dropout rate of 0.5. We set the batch size to 32. We use stochastic gradient descent over shuffled mini-batches with the Adadelta update rule \citep{zeiler2012adadelta} with $l_2$ constraint of 3. We experiment with a 5-fold cross-validation on the given source training set and report the average results.

\paragraph{Comparison Models}

We compare our models with the models from similar tasks as discussed in Section \ref{sec:competing}. Specifically, we compare with (a) Cross Domain Transfer (CDT) models \textsc{SCL} \citep{blitzer2007biographies} and \textsc{NeuSCL} \citep{ziser2017neural}, (b) CDT models with a CNN classifier integration \citep{yu2016learning} (\textsc{SCL+CNN} and \textsc{NeuSCL+CNN}), (c) a multiple-instance learning (MIL) model \textsc{MILNet} \citep{angelidis2018multiple}, (d) a weakly supervised model \textsc{MASM+CNN} \cite{amplayo2017aspect}. We remind that \textsc{MILNet} is only applicable to long to short transfer, and \textsc{MASM+CNN} is only applicable to short to long transfer. We use the available code provided by previous authors\footnote{\textsc{SCL} and \textsc{NeuSCL}: \url{https://github.com/yftah89/Neural-SCL-Domain-Adaptation}, \textsc{MILNet}: \url{https://github.com/stangelid/milnet-sent}, \textsc{MASM}: \url{https://github.com/rktamplayo/MicroASM}}. Finally, we also compare with \textsc{CNN} \citep{kim2014convolutional}, and a combination of two CNNs (\textsc{CNN}x2) as no-transfer baselines.

\paragraph{Dataset and Evaluation}

We use the datasets described in Table \ref{tab:data} for all our experiments. We use the following evaluation metrics. For all datasets, we use accuracy (\textsc{Acc}) to measure the overall sentiment classification performance. Additionally, for fine-grained datasets, we use root mean squared error (\textsc{RMSE}) to measure the divergence between the predicted and ground truth sentiment scores. Finally, in order to compare models in an integrated manner, we report the average transfer ratio \cite[TR;][]{glorot2011domain}, a version of the transfer loss which is more adaptive to averaging, calculated as the average quotient between the transfer error and the in-domain baseline error, i.e. $\text{TR} = \sum_x e(\mathcal{S}_x,\mathcal{T}_x) / e_b(\mathcal{T}_x,\mathcal{T}_x)$, where $\mathcal{S}_x$ and $\mathcal{T}_x$ are the source and domain of dataset $x$, respectively, $e$ and $e_b$ are accuracy errors from the competing model and the baseline \textsc{CNN} model.

\paragraph{Long to Short Transfer}
\label{sec:longshort}

We show the results for long to short transfer in the first part of Table \ref{tab:clt}. Results show that Cross Domain Transfer models do not perform well, which confirms our hypothesis that they are not well suited for this task. \textsc{MILNet} performs well on polarity tasks, but performs poorly on fine-grained tasks, having worse performance than the no-transfer CNN baseline. This shows that although Multiple Instance Learning models are effective in classifying positive or negative sentiments, they are not flexible to fine-grained sentiment intensities, which differs when text lengths are different. On the other hand, \textsc{BaggedCNN} performs better than \textsc{MILNet}, proving that simplifying MIL models work well on CLT. Overall, \textsc{LeTraNets} performs the best among all models, having the best accuracies and RMSEs on all datasets and settings.

\paragraph{Short to Long Transfer}

We report the results for short to long transfer in the second part of Table \ref{tab:clt}. Results show that Cross Domain Transfer models perform much worse compared to their performance in the long to short transfer task. The weak supervised model \textsc{MASM+CNN} performs the worst, having worse results than the no-transfer CNN baseline on all datasets. 
\textsc{BaggedCNN} also performs well in this task, even though it does not use its attention mechanism. This shows that \textsc{BaggedCNN} is a very tough-to-beat baseline for the CLT task.
Finally, \textsc{LeTraNets} also outperforms all the models on this subtask.

\paragraph{Transfer Ratio (TR)}

\begin{figure}[t]
    \centering
    \includegraphics[width=0.45\textwidth]{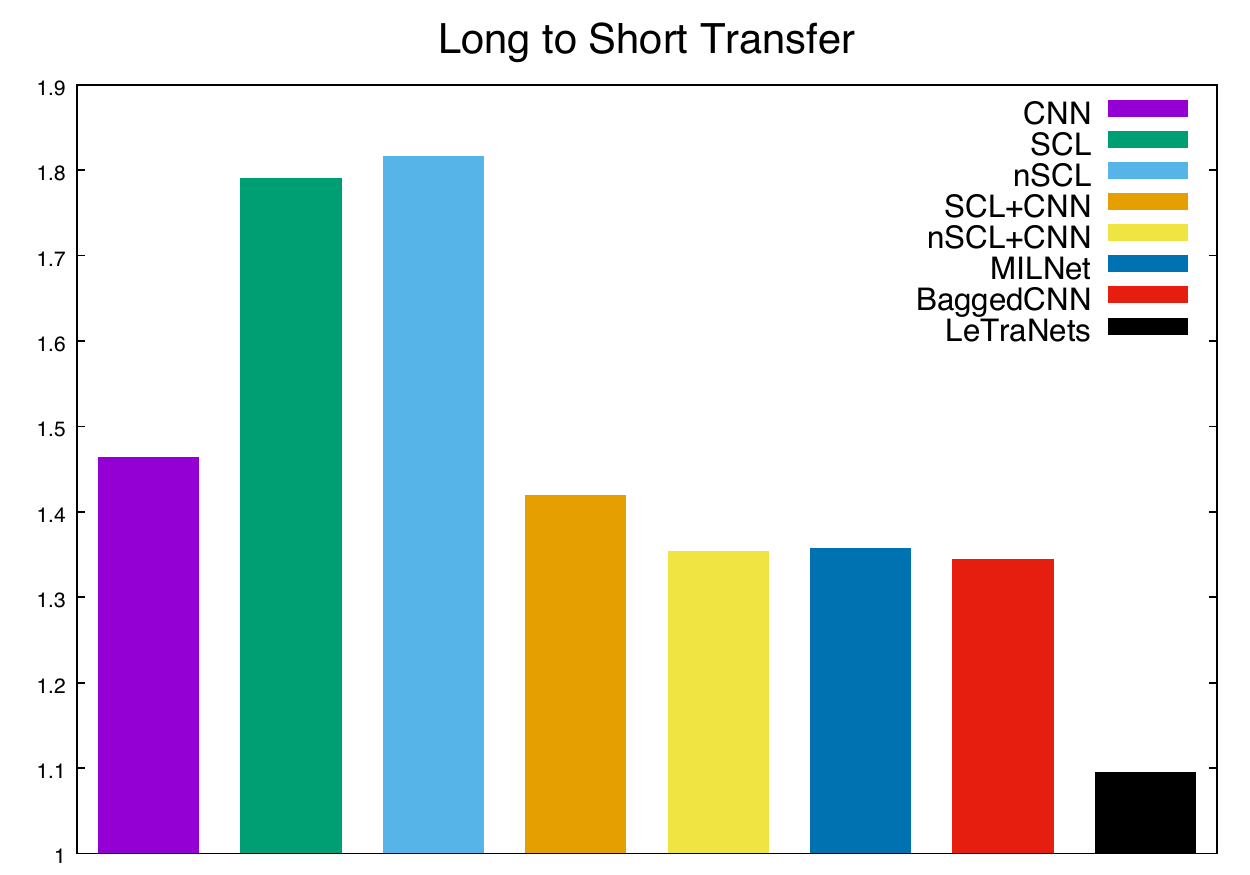}
    \includegraphics[width=0.45\textwidth]{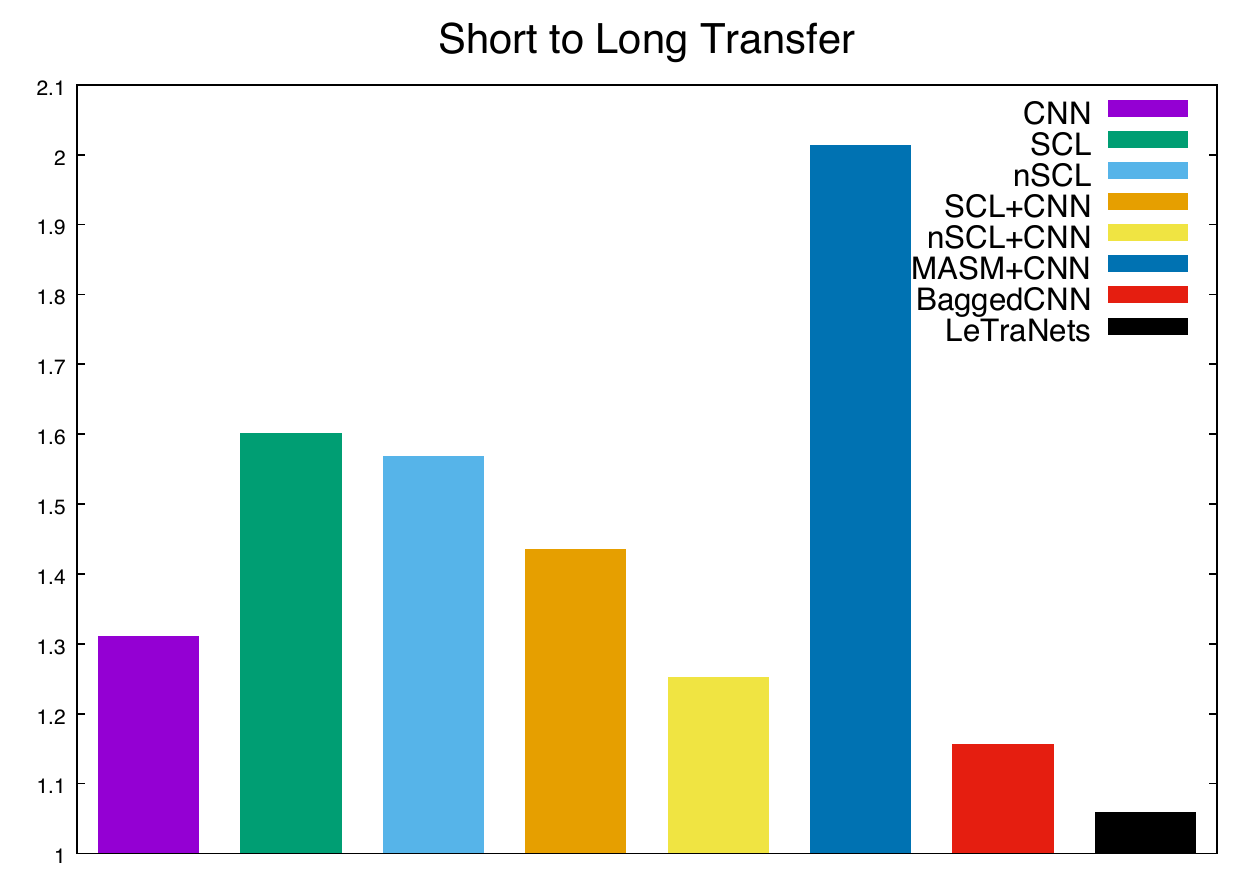}
    \caption{Average transfer ratio of all competing models for both transfer task. Lower is better.}
    \label{fig:transratio}    
\end{figure}

Figure \ref{fig:transratio} shows the average transfer ratio (TR) of all competing models, where $\text{TR}=1$ means trivially transferable.
The figure shows that the CDT models \textsc{SCL} and \textsc{NeuSCL} both obtain a larger transfer ratio compared to the no-transfer CNN baseline. The transfer ratios improve when CNN is integrated into both models, but does not improve much from the baseline. \textsc{MILNet} and \textsc{BaggedCNN} perform comparably on the long to short transfer task, where \textsc{BaggedCNN} performs slightly better.
\textsc{LeTraNets} performs the best among the models, having transfer ratios less than 1.1.

\begin{table}[t]
    \small
    \centering
    \begin{tabular}{cccccc}
        \thickhline
         & \textsc{Mov\_en}    & \multicolumn{2}{c}{\textsc{Res\_en}} & \multicolumn{2}{c}{\textsc{Mov\_ko}} \\
         & \textsc{P\_Acc}  & \textsc{P\_Acc}  & \textsc{F\_Acc} & \textsc{P\_Acc}  & \textsc{F\_Acc} \\
        \thickhline
        \multicolumn{6}{c}{\textit{Long to Short Transfer Task}} \\
        \hline
        - & 0.767  & 0.800  & 0.470 & 0.630  & 0.240 \\
        JT  & 0.782  & 0.838  & 0.480  & 0.637  & 0.255  \\
         PR  & 0.777  & 0.848  & 0.487  & 0.637  & 0.253  \\
         SP   & 0.782  & 0.845  & 0.495 & 0.635  & 0.260 \\
        All & \textbf{0.795} & \textbf{0.863} & \textbf{0.502} & \textbf{0.652} & \textbf{0.265} \\
        \thickhline
        \multicolumn{6}{c}{\textit{Short to Long Transfer Task}} \\
        \hline
        -  & 0.780  & 0.813  & 0.435  & 0.595  & 0.465 \\
        JT & 0.797  & 0.835  & 0.453 & 0.620  & 0.468 \\
        PR & \textbf{0.813} & 0.830  & 0.448 & 0.610  & 0.465 \\
        SP & 0.808  & 0.835  & 0.475  & 0.610  & 0.468  \\
        All & 0.810  & \textbf{0.858} & \textbf{0.478} & \textbf{0.625} & \textbf{0.493} \\
        \thickhline
    \end{tabular}%
  \caption{Effect (in \textsc{Acc}) of using the training mechanisms Joint Training (JT), Prediction Regularization (PR), and Stepwise Pretraining (SP) in \textsc{LeTraNets}.}
  \label{tab:effecttm}%
\end{table}

\section{Analyses}

\paragraph{Ablation on Training Mechanisms}

We investigate the performance of \textsc{LeTraNets} when the training mechanisms are not used. Specifically, we perform ablation tests on the Joint Training (JT), Prediction Regularization (PR), and Stepwise Pretraining (SP) mechanisms. The results in Table \ref{tab:effecttm} show that \textsc{LeTraNets} performs the best when all training mechanisms are used.
Also, when used individually, all the training mechanisms boost up the performance of the model. Hence, we confirm that the training mechanisms help \textsc{LeTraNets} achieve good performance on the task.

\paragraph{Performance per Text Length}

\begin{figure}[t]
    \centering
    \includegraphics[width=0.7\textwidth]{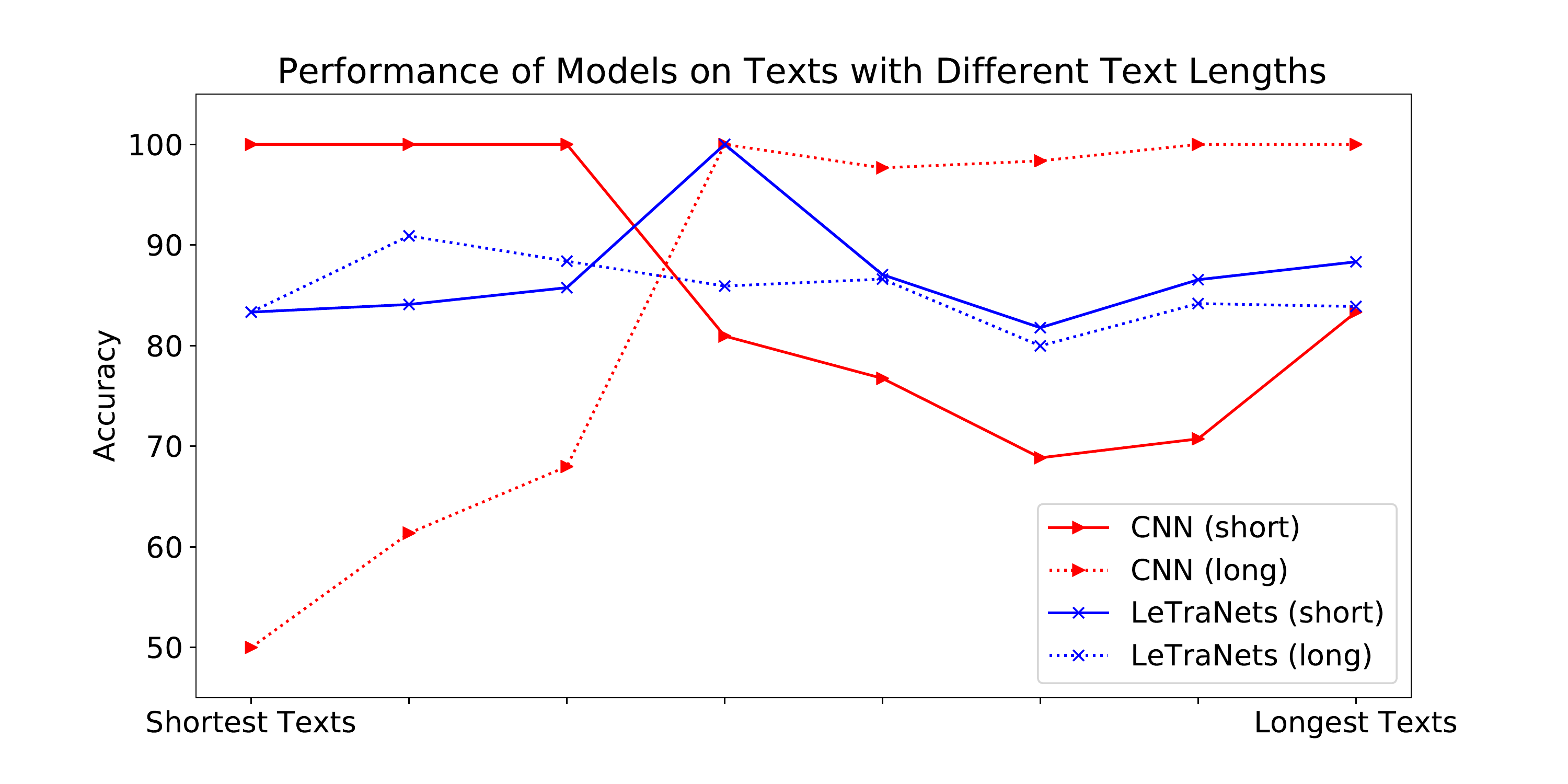}
    \caption{Accuracy of different models across different text lengths on the \textsc{Res\_en} dataset. The legend \texttt{model (length)} means that the \texttt{model} is trained on \texttt{length} dataset.}
    \label{fig:perlength}
\end{figure}

We check the capability of \textsc{LeTraNets} to transfer across text lengths, by looking at its performance as the text length increases.
Specifically, we compare the performance per text length of \textsc{LeTraNets} and CNN models, trained on either short texts (\textsc{LeTraNets}\textsubscript{short} and CNN\textsubscript{short}) or long texts (\textsc{LeTraNets}\textsubscript{long} and CNN\textsubscript{long}), on \textsc{Res\_en} short/long datasets.
Figure \ref{fig:perlength} shows the results.
CNN performs well when the text length is similar to the training dataset and performs poorly otherwise. \textsc{LeTraNets}, however, performs similarly on all kinds of text lengths although it is trained purely on a dataset of a specific length.
More interestingly, \textsc{LeTraNets}\textsubscript{short} performs \textit{better} than \textsc{LeTraNets}\textsubscript{long} on \underline{longer} texts, and unexpectedly performs \textit{worse} on \underline{shorter} texts. This suggests that \textsc{LeTraNets} weakens its ability to classify texts with the same length and improves its ability to classify texts with different length. 
This property is acceptable in our problem setup since we care on effectively classifying short (or long) texts more, assuming we only have access to long (or short) texts as training data.
However, future work should explore on CLT models that perform well on both text lengths.

\paragraph{On Topic Diversity}

\begin{figure}[t]
    \centering
    \includegraphics[width=0.7\textwidth]{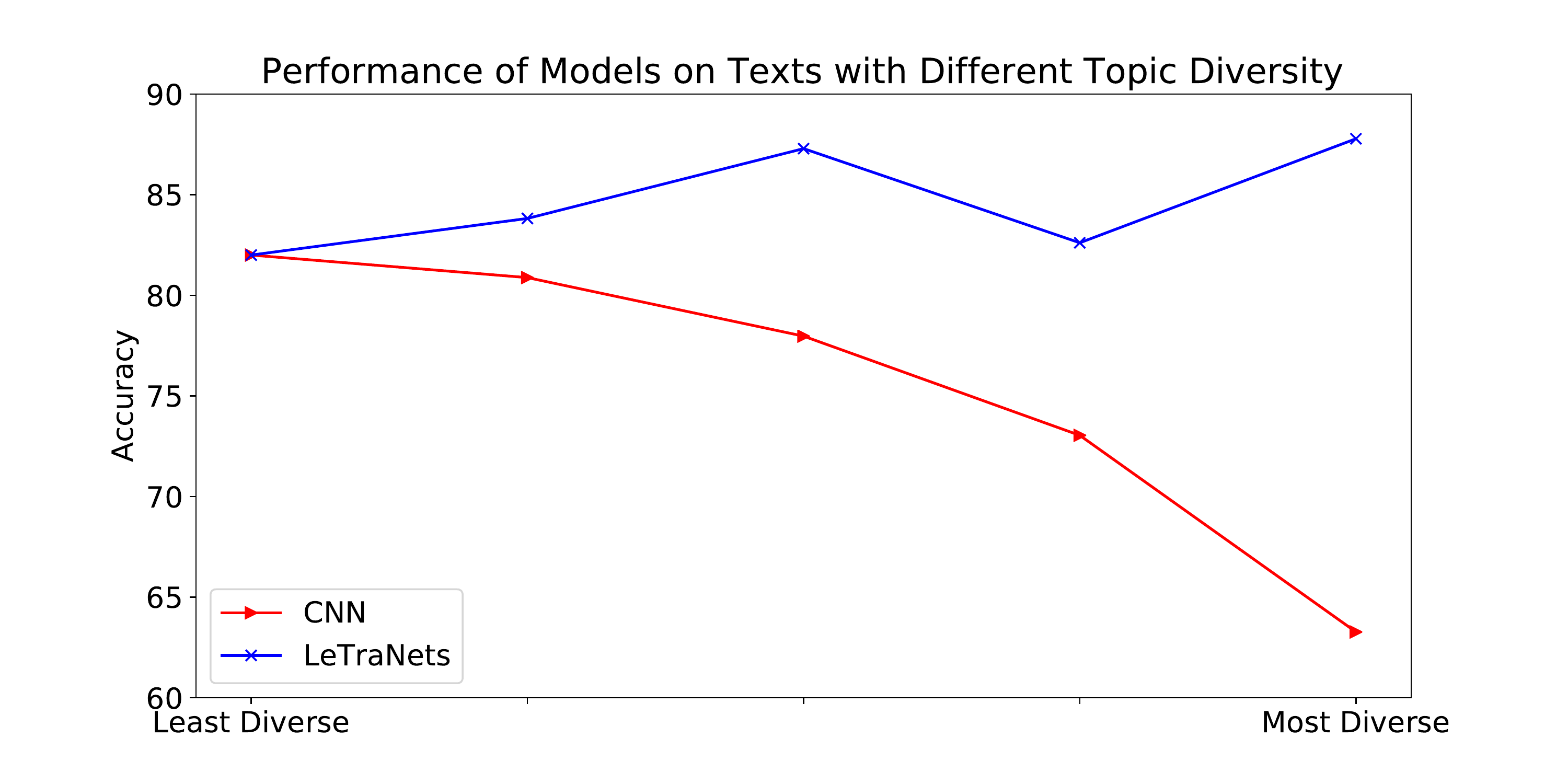}
    \caption{Accuracy of different models across different topic diversity on the \textsc{Res\_en} dataset.}
    \label{fig:pertopic}
\end{figure}

Longer texts can discuss diverse topics, while shorter texts are limited to few (or one) topics. 
In the sentiment classification domain, longer reviews may mention positive sentiments towards an aspect of a product, and then talk about negative sentiments towards another aspect.
With this hypothesis, we examine whether \textsc{LeTraNets} can handle longer texts with diverse topics when trained on short texts.
Specifically, we compare the performance per topic diversity of \textsc{LeTraNets} and \textsc{CNN} models, trained on short texts of \textsc{Res\_en} dataset.
We measure topic diversity as the Shannon index \citep{shannon1948mathematical} of the topic distribution inferred by an LDA topic model \citep{blei2003latent} fit using the unlabeled data.
Figure \ref{fig:pertopic} shows the results.
Results indicate that the performance increase of \textsc{LeTraNets} over \textsc{CNN} increases as the diversity of topics increases.
This shows that for short to long transfer, \textsc{LeTraNets} is able to handle texts with topics that are more diverse, even when trained on short texts, which tend to have less diverse topics.

\paragraph{Cross Domain and Length Transfer}

Which between domain and text length should we consider to achieve a better performance? To answer this question, we combine Cross Domain Transfer (CDT) and Cross Length Transfer (CLT) into one task: Cross Domain and Length Transfer (CDLT) and compare the performance of CDT and CLT models on the task. We use the \textsc{Mov\_en} and \textsc{Res\_en} datasets to create four CDLT datasets, and check which between the CDT model \textsc{NeuSCL+CNN} and the CLT model \textsc{LeTraNets} achieves a higher increase in performance. The results are shown in Table \ref{tab:domainlength}. We find that \textsc{NeuSCL+CNN} performs worse, obtaining accuracies worse than that of the no-transfer CNN baseline.
\textsc{LeTraNets} performs better, obtaining significant increase in performance from the baseline. This shows that solving the non-transferability of length is more important to achieve a more effective sentiment classifier.

% Table generated by Excel2LaTeX from sheet 'Sheet6'
\begin{table}[t]
  \small
  \centering
    \begin{tabular}{lcccc}
    \thickhline
     & \multicolumn{2}{c}{Long to Short} & \multicolumn{2}{c}{Short to Long}\\
    \multicolumn{1}{c}{Model} & M\textgreater R    & R\textgreater M    & M\textgreater R    & R\textgreater M \\
    \hline
    \textsc{CNN} & {0.755} & {0.620}  & {0.755} & {0.728} \\
    \textsc{NeuSCL+CNN} & \textcolor{red}{0.710} & {0.638} & \textcolor{red}{0.695} & \textcolor{red}{0.585}  \\
    \textbf{\textsc{LeTraNets}} & {\textbf{0.828}} & {\textbf{0.745}} & {\textbf{0.790}} & {\textbf{0.785}} \\
    \thickhline
    \end{tabular}%
  \caption{Accuracies of models on four CDLT datasets from M:\textsc{Mov\_en} and R:\textsc{Res\_en} datasets.}
  \label{tab:domainlength}%
\end{table}%

%\paragraph{Error Analysis}

%We look into the errors produced by \textsc{LeTraNets} and find two patterns that stand out.
%First, error instances contain many out-of-vocabulary (OOV) words, averaging to 23.6 words, which constitutes to almost 10\% of the text.
%Examples of OOV are wrongly spelled words, company names, and Spanish words. This means there are differences in word and entity usage, despite both being from the same domain. In our prior experiments, we tried methods such as Similarity Regularization \citep{ziser2017neural} and pretrained word vectors \citep{chen2016neural}. However, both provide no gains in performance.

%Second, 98.3\% of the error instances contain words that aid transition between two segments, such as transition and conjunction words. The average number of conjunctions in a text is 11.4. The conjunctions\footnote{We use the list of conjunctions provided here: \url{http://www.smart-words.org/linking-words/conjunctions.html}} that appear the most are the ``but'' (17.6\%) conjunctions, the ``and'' (13.7\%) and ``or'' (11.3\%) conjunctions, and the ``for'' (11.5\%) and ``so'' (8.2\%) conjunctions.
%The literature suggests the usage of RNNs \citep{sutskever2011generating} such as LSTMs to model inter-sentence transitions in classification \citep{yang2016hierarchical}, however we show in Section \ref{sec:longshort} that incorporating RNNs, such as in MILNet \citep{angelidis2018multiple}, hurts the performance of the model.

\section{Conclusions}

We defined a new task called Cross Length Transfer (CLT) to check the transferability across lengths of classification models. We set the grounds by defining the task, providing three benchmark datasets from different domains and languages, and introducing models from related tasks. We proposed two models: a strong baseline model called \textsc{BaggedCNN}, and \textsc{LeTraNets}, a model that improves over the weakness of \textsc{BaggedCNN}. Our multiple experiments show that \textsc{LeTraNets} demonstrates superior performance over all competing models. 
We aim to apply the CLT to other classification tasks, such as natural language inference \citep{bowman2015large}, where text length is influential towards overall model performance \citep{williams2017broad}.

\section*{Acknowledgments}

We would like to thank the anonymous reviewers for their helpful feedback and suggestions. Amplayo is grateful to be supported by a Google PhD Fellowship. This research was supported by MSIT (Ministry of Science and ICT), Korea, under ITRC program (IITP-2019-2016-0-00464) supervised by IITP. Hwang is the corresponding author.

\bibliography{acml19}

\end{document}